\documentclass[letterpaper, 10 pt, conference]{ieeeconf}  

\IEEEoverridecommandlockouts                              

\usepackage{graphics} 
\usepackage{epsfig} 
\usepackage{times} 
\usepackage{amsmath} 
\usepackage{amssymb}  
\usepackage{subcaption}

\author{\IEEEauthorblockN{Teppei Suzuki}
\IEEEauthorblockA{\textit{Denso IT Laboratory, Inc.} \\
Tokyo, Japan \\
suzuki.teppei@core.d-itlab.co.jp}
}

\title{\LARGE \bf
Federated Learning for Large-Scale Scene Modeling\\with Neural Radiance Fields
}

\author{Teppei Suzuki$^{1}$
\thanks{$^{1}$Denso IT Laboratory, Inc., Tokyo, Japan}
\thanks{{\tt\small suzuki.teppei@core.d-itlab.co.jp}}}%

\begin{document}

\maketitle
\thispagestyle{empty}
\pagestyle{empty}

\begin{abstract}
We envision a system to continuously build and maintain a map based on \textit{earth-scale} neural radiance fields (NeRF) using data collected from vehicles and drones in a lifelong learning manner.
However, existing large-scale modeling by NeRF has problems in terms of scalability and maintainability when modeling earth-scale environments.
Therefore, to address these problems, we propose a federated learning pipeline for large-scale modeling with NeRF.
We tailor the model aggregation pipeline in federated learning for NeRF, thereby allowing local updates of NeRF.
In the aggregation step, the accuracy of the clients' global pose is critical.
Thus, we also propose global pose alignment to align the noisy global pose of clients before the aggregation step.
In experiments, we show the effectiveness of the proposed pose alignment and the federated learning pipeline on the large-scale scene dataset, Mill19.
\end{abstract}

\section{Introduction}
Neural radiance fields (NeRF)~\cite{nerf} have emerged to represent scenes for view synthesis and have been used as maps for many robotics applications, such as vision-based localization, navigation, and SLAM~\cite{inerf,imap,nice-slam,adamkiewicz2022vision,kwon2023renderable,loc-nerf}.
Although these studies have been evaluated in relatively small-scale scenes, we believe that such NeRF-based robotics applications will be available to the systems operated in large-scale environments, such as self-driving cars and delivery drones.
In addition, since NeRF can lift arbitrary 2D information to 3D (\textit{e.g.}, semantic labels~\cite{kundu2022panoptic}, and feature vectors~\cite{kobayashi2022decomposing}), it will facilitate a higher-level environmental recognition and understanding in robotics applications.

Nowadays, a large number of vehicles are on the road, and in the near future, many unmanned aerial vehicles, such as delivery drones, will be in the sky.
In such a world, we envision the system to continuously build and maintain an \textit{earth-scale} NeRF-based map using data collected from vehicles and drones in a life-long learning manner and to leverage the map created and maintained with such a system.

Large-scale scene modeling with NeRF has been studied in recent years~\cite{block-nerf,mega-nerf}.
These methods divide a large-scale environment into many small areas and model the small areas with multiple NeRF models.
Although this pipeline efficiently models a large-scale scene in a distributed training manner, it has three problems in realizing our envisioned system: (i) The existing pipelines need to collect data in a server once, but they need high communication costs to collect data from clients on a server and need a large number of memory budgets to save the earth-scale data; (ii) these pipelines require tremendously large computational resources for training a large number of models to cover an earth-scale environment; and (iii) they can only update the model for each pre-divided area for the maintenance of the NeRF-based map, and if the training data is collected only in a part of the area, the model may suffer from the forgetting problem~\cite{imap}.

\begin{figure}
    \centering
    \includegraphics[clip,width=1\hsize]{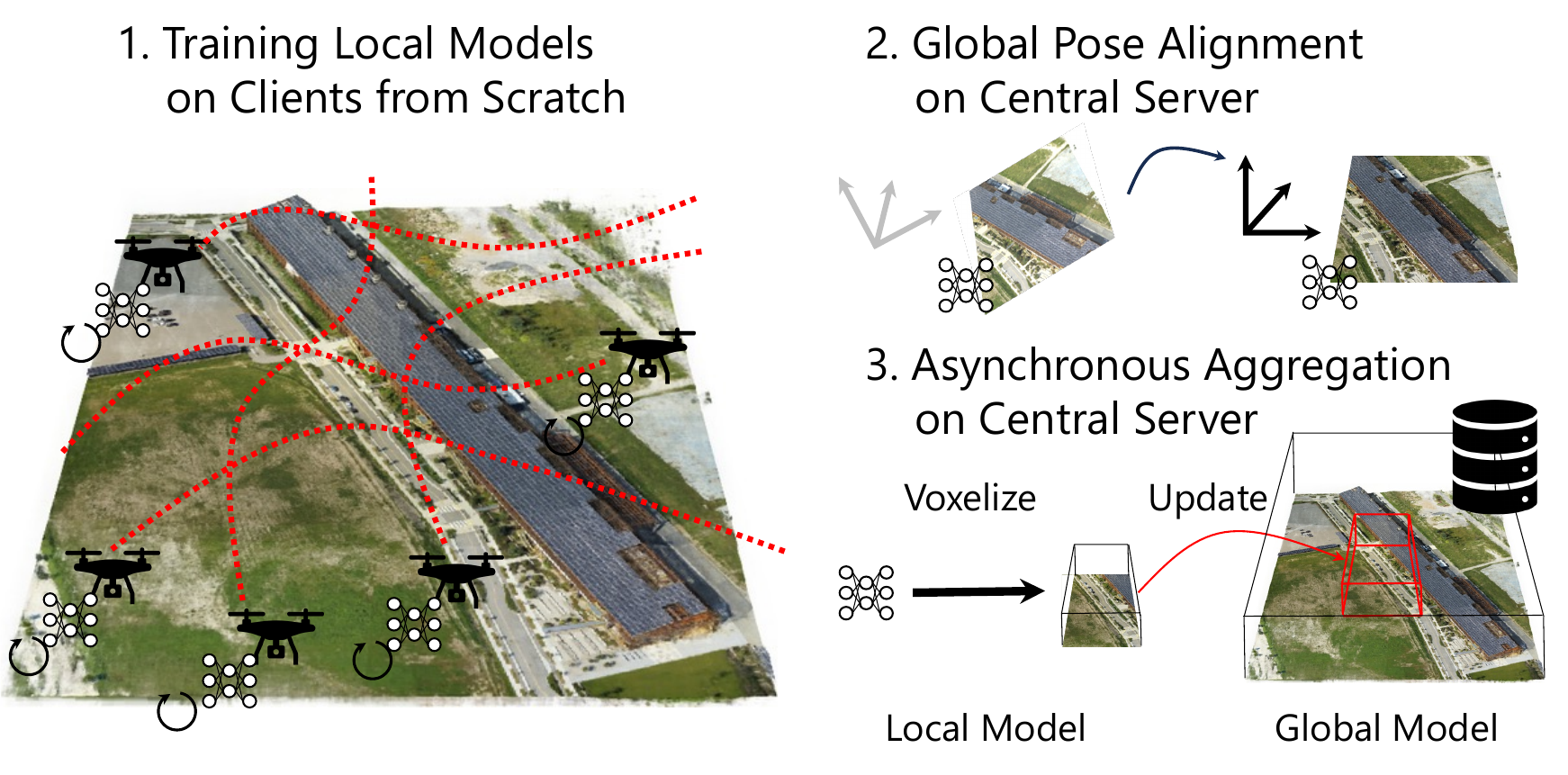}
    \caption{Overview of the proposed federated learning pipeline for neural radiance fields.}
    \label{fig:overall}
\end{figure}

To develop a more scalable and maintainable method, we propose the federated learning pipeline for large-scale modeling with NeRF, as depicted in Fig. \ref{fig:overall}.
Federated learning (FL)~\cite{fed-learn} is a data-decentralizing training protocol that enables clients, such as mobile devices, to collaboratively train a model while keeping the training data decentralized.
Since FL does not need to aggregate data collected by clients on the server, it can alleviate the communication cost problem and the data storage problem.
FL also alleviates the computational resource problem because it can leverage the resources of clients (millions or more in our scenario, and then their overall FLOPS will be much larger than those of supercomputers).

To train NeRF in a federated learning manner, we tailor the model aggregation pipeline, which allows the global model to be locally updated based on the client data, avoiding the forgetting problem and improving maintainability.
Specifically, we cache the outputs of each local model as the 3D voxel grid, and then the global model is updated by aggregating cached representation based on the clients' pose in the global coordinate.
This pipeline does not affect the outputs of the global model beyond the region modeled by the local models.
In addition, to align the inaccurate global pose of clients due to sensor noise, we propose the global pose alignment step before the aggregation step, which is inspired by the vision-based localization using NeRF~\cite{inerf,loc-nerf,lin2023parallel}.

Our contribution is summarized below:
\begin{itemize}
    \item We propose the federated learning pipeline for large-scale scene modeling with NeRF. By caching the outputs of each local model as the voxel grid, we realize the local update of the global model.
    \item We also propose the global pose alignment. To alleviate the sensor noise in the clients' pose in the global coordinate, we align the global pose of each client before the aggregation step by minimizing the difference between the RGB and depth images rendered by the global and local models.
    \item We assess the effectiveness of our training pipeline on the Mill19 dataset~\cite{mega-nerf}, which contains thousands of HD images collected from drone footage over 100,000m$^2$ of terrain near an industrial complex.
\end{itemize}

\section{Related Work}
\subsection{Neural Radiance Fields}
3D reconstruction from image collections~\cite{schoenberger2016sfm,agarwal2011building,pollefeys2008detailed} is an important task in computer vision fields, and the reconstruction methods have been applied to many robotics applications, particularly SLAM systems~\cite{mur2015orb,teed2021droid,engel2014lsd}.
In principle, objects or whole scenes are reconstructed as 3D points based on multiple-view geometry.

In recent years, NeRF~\cite{nerf} has emerged for novel view synthesis.
Unlike 3D point-based reconstruction, NeRF models the target scene by a continuous function and its image rendering process is differentiable.
Taking advantage of these properties, some vision-based robotics systems have been proposed.
iNeRF~\cite{inerf}, a vision-based localization method, estimates the camera pose through the minimization of the loss between the target image and the rendered image with the gradient method.
Maggio \textit{et al}.~\cite{loc-nerf} and Lin \textit{et al.}~\cite{lin2023parallel} improved the robustness of iNeRF by leveraging the advantage of Monte Carlo localization~\cite{dellaert1999monte}.
Adamkiewicz \textit{et al}.~\cite{adamkiewicz2022vision} utilized this localization scheme for robot navigation.
Kruzhkov \textit{et al}.~\cite{kruzhkov2022meslam} incorporated NeRF into the SLAM system.
They used NeRF as a mapping module and reduced the memory budget for the map.
Moreover, pure NeRF-based SLAM systems, such as iMap~\cite{imap} and Nice-SLAM~\cite{nice-slam}, were studied.

As reviewed above, NeRF is applied to robotics applications that use NeRF as a map.
When considering systems using NeRF as a map in a large environment, such as self-driving cars and delivery drones, we need to model large-scale scenes.
There are certain studies for large-scale scene modeling by NeRF.
Block-NeRF~\cite{block-nerf} and Mega-NeRF~\cite{mega-nerf} have modeled urban-scale environments.
Both methods divide the target scene into small areas and then model the small areas with multiple NeRF models.
They can efficiently model a large-scale scene in a distributed training manner.

However, the existing large-scale modeling methods have some problems in terms of maintainability and computational costs.
When part of the scene is changed, they need to update all models representing the changed scene, even if the changed area is a part of the modeled area.
In addition, due to the forgetting problem~\cite{imap}, the model needs to be trained with all data in the corresponding area if a part of the area is changed.
Therefore, maintaining the map created by existing methods is costly.
Moreover, these methods require tremendously large memory budgets to save data and large computational resources to model the environment if one builds an earth-scale NeRF-based map.
Hence, we approach these problems by adopting a federated learning scheme.

\subsection{Federated Learning}
Federated learning (FL)~\cite{fed-learn} is a data-decentralizing training process for machine learning models.
It distributes the model to clients such as mobile devices or organizations, trains the model on clients using their own data, and aggregates the models on the server.
By repeating the above process, FL collaboratively trains a global model under the orchestration of a central server.
FL has some advantages: preserving privacy since it does not collect user data on the server and leveraging the computational resources of many clients.
Because of these advantages, FL has been studied for training machine learning models in various tasks, such as image classification~\cite{fed-learn,8761315}, image segmentation~\cite{miao2023fedseg}, and self-supervised learning~\cite{fedema,zhang2020federated}.

Many FL approaches update models using a synchronous protocol, which needs to wait for all updates on clients before aggregation and delays training.
The existence of lagging devices (i.e., stragglers, stale workers) is inevitable due to device heterogeneity and network unreliability.

To mitigate this problem, asynchronous federated learning (AFL) has been proposed~\cite{xie2019asynchronous}, which aggregates local models asynchronously.
Xie \textit{et al.}~\cite{xie2019asynchronous} aggregates models by exponential moving averaging.
Chen \textit{et al.}~\cite{chen2020asynchronous} proposed a method to aggregate the model with a decay coefficient to balance the previous and current gradients on clients.
Since AFL improves the communication efficiency of FL, our pipeline also adopts the AFL strategy.

\section{Federated Learning for NeRF}
We present the overall pipeline of our method in Fig. \ref{fig:overall}, which consists of three steps, training local models, global pose alignment, and an aggregation step.
We consider the AFL setting; namely, the server updates the global model as soon as it receives the local model.

We first describe the assumptions of this study in Sec. \ref{sec:assumption}.
Then, in Secs. \ref{sec:local} and \ref{sec:global}, we describe the training pipeline for the local models and the aggregation step.
Finally, we describe the global pose alignment in Sec. \ref{sec:pose-align}.

\subsection{Problem Definition and Assumptions}
\label{sec:assumption}
In 3D reconstruction with federated learning, there are several challenges due to keeping data decentralized.
As major concerns, (i) the local models trained by clients are not build on the shared global coordinate.
In other words, rendered images from the same camera through different local models may show different regions.
(ii) the local models may have different appearance because each client may collect data at different time (\textit{e.g.}, nighttime and daytime).
In this work, we approach the former problem, and the latter problem is left as future work; namely, we assume that images collected by clients are captured at the same time.

We assume each client knows the relative camera poses between collected images.
This is a realistic assumption because relative poses can be obtained from some sensors, SLAM, or SfM.
In the experiments, we use the camera poses given by Mill19~\cite{mega-nerf} that are estimated by PixSfM~\cite{lindenberger2021pixel}.
Note that if the obtained poses are noisy, we would be able to correct them by using the methods that simultaneously optimize poses and the model (\textit{e.g.}, BARF~\cite{barf} and NoPe-NeRF~\cite{nope-nerf}).

We also assume that the global pose of each client is known but \textit{noisy}.
Since the goal of this study is to train the large-scale NeRF model as a map, there is no map to localize the position\footnote{Once the map (\textit{i.e.}, the global model) is initialized, we can use it to localize each client in the same manner as that for iNeRF~\cite{inerf}.}.
In this situation, we need to use sensors such as GPS and IMU to obtain the global pose, but the pose obtained from the sensors basically contains noise.

\subsection{Training Local Models}
\label{sec:local}
For simplifying the pipeline, we assume that client models are trained from scratch because NeRF basically ``overfits'' training data and it does not care how the model is initialized.
Of course, we can consider the standard pipeline; namely, clients download the global model and then train it with the local data, which may make convergence faster but increase communication costs.

The local model consists of two models: the density model $f_\sigma$ and the color model $f_c$.
Let $\mathbf{r}=\{\mathbf{r}_i\in\mathbb{R}^3|\mathbf{r}_i=\mathbf{o}+s_i\mathbf{d},\ s_i\in\mathbb{R}\}_i$ be a set of sampled points along the ray passing through a pixel, where $\mathbf{o}$ and $\mathbf{d}$ are a camera origin and a ray direction.
Following \cite{mega-nerf}, we sample 512 points per ray for rendering images in our experiments.
The density at $\mathbf{r}_i$ is computed as $\sigma_i=\phi\circ f_\sigma(\mathbf{r}_i)$, where $\phi$ denotes the softplus function.
For representing RGB, we use the spherical harmonics (SH) as in PlenOctrees~\cite{plenoctrees}, which enables caching outputs of the model on the grid voxels in the aggregation step described in Sec. \ref{sec:global}.
Specifically, $f_c$ outputs SH coefficients $\mathbf{k}(\mathbf{r}_i)=f_c(\mathbf{r}_i)$, where $\mathbf{k}(\mathbf{r}_i)=\{k_\ell^m\in\mathbb{R}^3\}_{\ell:0\leq\ell\leq\ell_{\max}}^{m:-\ell\leq m\leq\ell}$; $\ell$ and $m$ are a degree and an order of the SH function, respectively.
Each $k_\ell^m$ is a set of three coefficients corresponding to RGB.
Then, the view-dependent color at $r_i$ is computed by
\begin{align}
    \label{eq:color}
    \mathbf{c}_i=S\left(\sum_{\ell=0}^{\ell_{\max}}\sum_{m=-\ell}^{\ell}k_\ell^mY_\ell^m(\mathbf{d})\right),
\end{align}
where $S(\cdot)$ is the sigmoid function and $Y_l^m(\mathbf{d}):\mathbb{S}^2\rightarrow\mathbb{R}$ is the SH function at the viewing angle $\mathbf{d}$.
Following \cite{plenoxels}, we use spherical harmonics of degree 2, which has 27 harmonic coefficients.

The rendering and training procedure is the same as in the original NeRF~\cite{nerf}.
Specifically, the pixel color corresponding to the ray $\mathbf{r}$ is computed as follows:
\begin{align}
    \label{eq:rendering}
    \hat{\mathbf{C}}(\mathbf{r})=\sum_i\exp(-\sum_{j=1}^{i-1}\sigma_j\delta_j)(1-\exp(-\sigma_i\delta_i))c_i,
\end{align}
where $\delta_i=t_{i+1}-t_i$ is the distance between adjacent samples on the ray.
Then, the NeRF model is trained by minimizing the following loss:
\begin{align}
    \label{eq:loss}
    \mathcal{L}=\underset{\mathbf{r}\in\mathcal{R}}{\mathbb{E}}\left\|\hat{C}(\mathbf{r})-C(\mathbf{r})\right\|^2_2,
\end{align}
where $C(\mathbf{r})\in\mathbb{R}^3$ denotes the ground-truth pixel color corresponding to $\mathbf{r}$, and $\mathcal{R}$ is the set of rays in the training data.
Note that we do not consider the coarse-to-fine rendering pipeline~\cite{nerf} in this study, but it can be made available with a larger memory budget and more computational resources.

\subsection{Updating Global Model}
\label{sec:global}
Averaging parameters, often used in federated learning~\cite{fed-learn}, cannot work for NeRF due to its properties;
NeRF represents only the scene in the training data and assigns random color and density to the outside of the scene.
In addition, the 3D position $(x,y,z)$ may indicate a different location for each client because clients' coordinates may not be aligned due to the sensor noise.
For these reasons, the simple averaging strategy will degrade the rendering quality of the global model.

\begin{figure}
    \centering
    \includegraphics[clip,width=0.9\hsize]{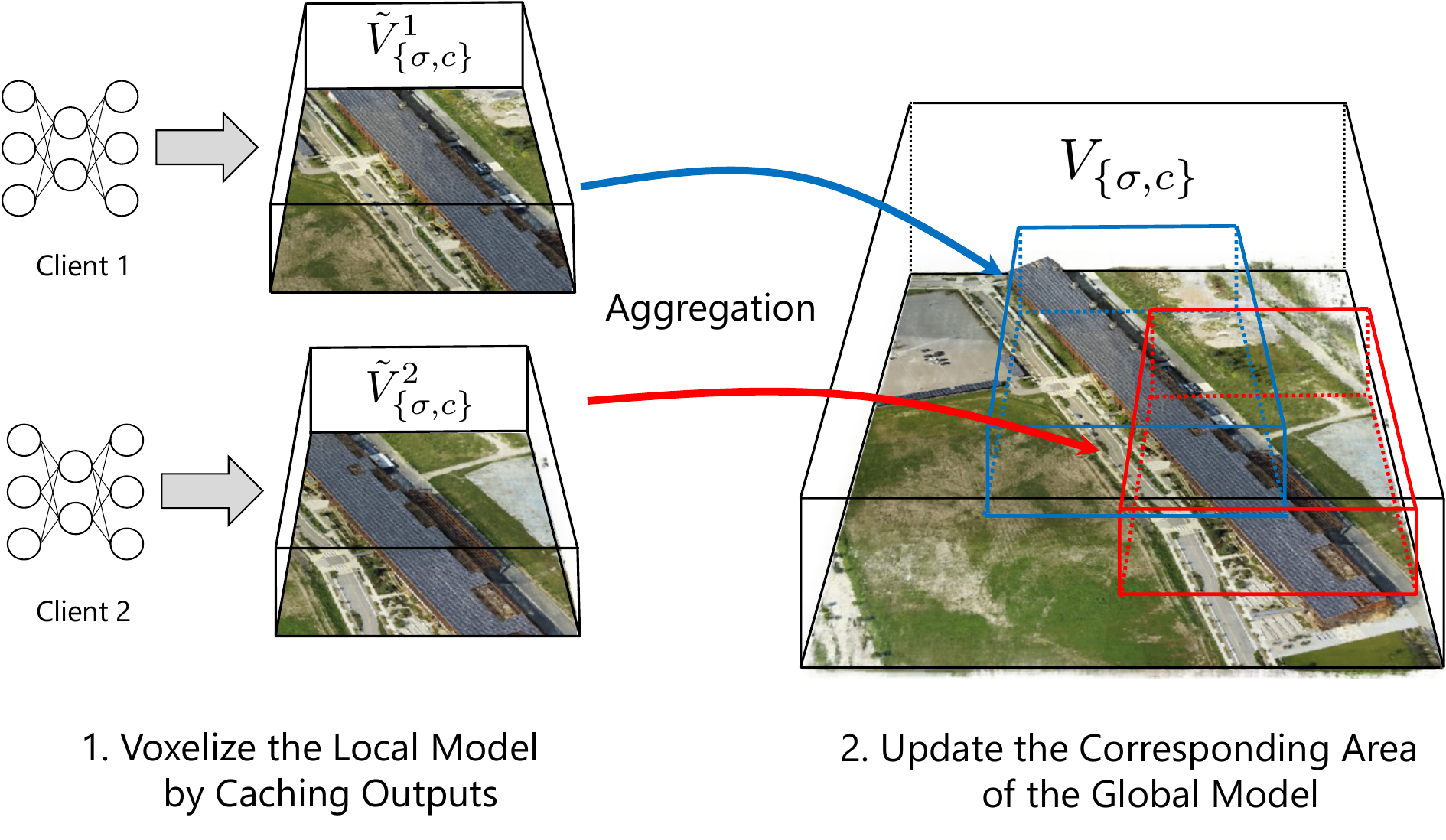}
    \caption{Overview of the proposed aggregation step. We first cache outputs of the local model on the grid voxels, $\tilde{V}^i_\sigma$ and $\tilde{V}^i_c$, and then add them to the global voxel grid, $V_\sigma$ and $V_c$.}
    \label{fig:aggregation}
\end{figure}

To avoid the aforementioned problem, we tailor the aggregation step for NeRF.
We represent the global model as the voxel grid, as in Plenoxels~\cite{plenoctrees}.
We cache the output of the local models on the voxel grid, as in \cite{fastnerf,sekikawa2019tabulated,sekikawa2020irregularly}, and then aggregate the cached voxel grid and the global model, as shown in Fig. \ref{fig:aggregation}.
Since we only cache the outputs in the area modeled by the local model and aggregate them, this procedure does not affect the output outside the area.
In addition, the voxel grid representation allows us to align the global pose of the clients before aggregation  (Sec. \ref{sec:pose-align}).
For other advantage of our aggregation scheme, it does not depend on the architecture of the local model, which is important for federated learning because we can select an appropriate architecture based on the client's computational resources.

Let $\tilde{V}_\sigma^n\in\mathbb{R}^{X\times Y\times Z\times 1}$ and $\tilde{V}_c^n\in\mathbb{R}^{X\times Y\times Z\times 27}$ be a cached density grid voxel and a cached SH coefficient grid voxel of the $n$-th client, where $X$, $Y$, and $Z$ are the voxel resolution and $27$ in $V_c$ corresponds to the number of the coefficients.
Also, we define the global model's voxel grids as $V_\sigma\in\mathbb{R}^{X_g\times Y_g\times Z_g\times 1}$ and $V_c\in\mathbb{R}^{X_g\times Y_g\times Z_g\times 27}$, where  $X_g$, $Y_g$, and $Z_g$ are the global model's voxel resolution.
Note that the global model is defined on the global coordinate and the local voxel grid is defined on the local coordinate, respectively; hence, the position on the local coordinate, $P\in\mathbb{R}^3$, is mapped to the global coordinate using the global pose, as in $P_g=R_nP+t_n$, where $R_n\in \mathrm{SO}(3)$ and $t_n\in\mathbb{R}^3$ denote the relative rotation and translation between the $n$-th client's local and global coordinates, respectively, which are obtained from sensors such as GPS and IMU.
Then, we update the density grid voxel of the global model with exponential moving averaging in the asynchronous federated learning manner as
\begin{align}
    \label{eq:ema}
    V_{\sigma}[P_g]\leftarrow\eta V_{\sigma}[P_g] + (1-\eta)\tilde{V}_{\sigma}^n[P],
\end{align}
where $V_{\sigma}[P]$ denotes an element of $V_{\sigma}$ at $P$; $\eta$ denotes a mixing value that is set to 0.9 in the experiments.
The SH coefficient grid is also updated in the same manner.
Note that $\tilde{V}_{\sigma}^n[P]$ is the output of the local model (\textit{i.e.}, $\phi\circ f_\sigma(P)$), and we directly add $\phi\circ f_\sigma(P)$ to $V_{\sigma}[P_g]$, instead of $\tilde{V}_{\sigma}^n[P]$.
Therefore, we can ignore correspondence between nodes of the grids, $V_{\sigma}$ and $\tilde{V}$.

Compared to directly replacing the parameters of the global model with those of the local model, exponential moving averaging would mitigate unreliable outputs of the local model.
Since NeRFs model scenes from multiple view images without geometry constraints, they sometimes represent scenes with incorrect 3D shapes, and it causes degradation of the novel view synthesis quality.
However, in exponential moving averaging, the effect of a single client on the global model is negligible when $\eta$ is small.
Thus, the noisy results would be ignored if clients can model scenes well on average.

The rendering process of the global model is the same as Plenoxels~\cite{plenoxels}; namely, the density and SH coefficient are sampled from the voxel grids by trilinear interpolation and then the RGB color is computed following eq. \eqref{eq:color} with the sampled coefficients.
Finally, pixel colors are rendered by eq. \eqref{eq:rendering} with the sampled density and the color.
We define the rendered pixel color through the global model as $\hat{C}_g(\mathbf{r})$.

\subsection{Global Pose Alignment\label{sec:pose-align}}
\label{sec:pose-align}
As we mentioned in Sec. \ref{sec:assumption}, the global pose of the clients contains noises.
Thus, we need to correct it before aggregation because inconsistency between the local and global models occurs in eq. \eqref{eq:ema} if the global pose is incorrect.

To align the pose, we optimize it to minimize the difference between the reference and target images, which are rendered by the local and global models, respectively.
Let $t^\ast\in\mathbb{R}^3$ and $\{R_j^\ast\in \mathrm{SO}(3)\}_j$ be a translation vector and rotation matrices for the target views, which are obtained from the area modeled by both the global and local models.
Then, we align the client's global pose by solving the following minimization problem:
\begin{align}
    \nonumber
    (\hat{t},\ \hat{R})=\underset{\{t,R\}}{\arg\min} \sum_j&\underset{\mathbf{r}_{t^\ast,R_j^\ast}}{\mathbb{E}}\lambda|\hat{C}_g(\mathbf{r}_{t^\ast,R_j^\ast})-\hat{C}(\mathbf{r}_{t^\ast+t,RR_j^\ast})|\\
    \label{eq:pose-align}
    +(1-\lambda)&|\hat{D}_g(\mathbf{r}_{t^\ast,R_j^\ast})-\hat{D}(\mathbf{r}_{t^\ast+t,RR_j^\ast})|,
\end{align}
where $\mathbf{r}_{t, R}$ denotes the ray depending on the camera pose $(t, R)$, and $\hat{D}_g$ and $\hat{D}$ denote the rendered depth through the global and local models, respectively, which are computed by replacing color values, $c_i$, in eq. \eqref{eq:rendering} with the sampled point position, $s_i$, as in \cite{imap}; $\lambda\in[0,1]$ is a weight parameter.
After optimization, we align the global pose of the local model based on $\hat{t}$ and $\hat{R}$.
Since the rendering process can be differentiable with respect to $t$ and $R$, we optimize them using the gradient method, as in the existing NeRF-based localization methods~\cite{inerf,loc-nerf,lin2023parallel}.
In our experiments, we used the same optimization pipeline as that in Lin \textit{et al}.~\cite{lin2023parallel} to optimize the pose, which is a Monte Carlo-based method.
Note that, unlike existing methods~\cite{inerf,loc-nerf,lin2023parallel}, we leverage multiple views and rendered depth to make optimization stable.
We assess the effectiveness of the multiple target views and the depth loss in the experiment section.

\begin{figure*}
    \centering
    \begin{tabular}{cc}
        \includegraphics[clip,width=0.45\hsize]{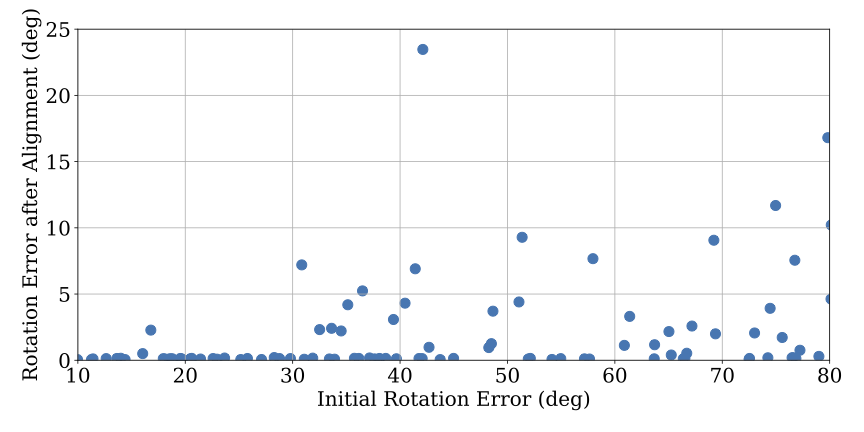} &  
        \includegraphics[clip,width=0.45\hsize]{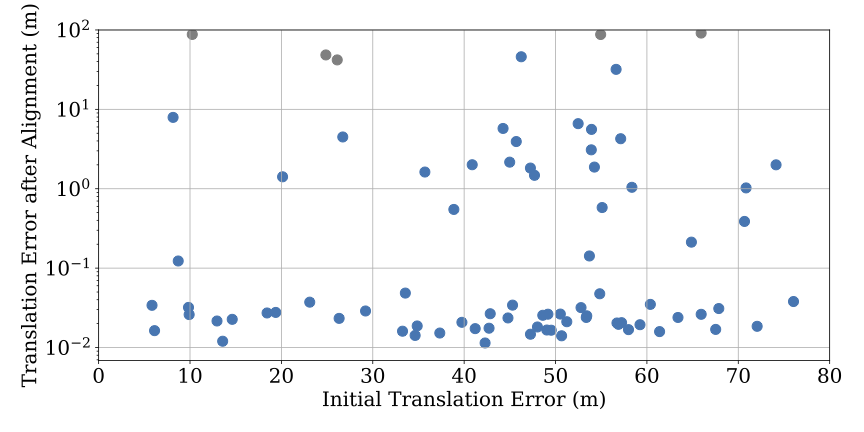}
    \end{tabular}
    \caption{The alignment results for various initial errors. The gray dots denote failure cases that increase the initial errors.}
    \label{fig:var-noise}
\end{figure*}
\section{Experiments}
\subsection{Experimental Setup}
\label{sec:setup}
To simulate our method, we use the Mill19 dataset~\cite{mega-nerf} that includes two scenes, \texttt{building} and \texttt{rubble}, which have thousands of HD images collected from drone footage over 100,000m$^2$ of terrain close to an industrial complex.
We first generate the data owned by clients in the following procedure: first, we randomly select one image from training data and compute the Euclidean distance between the camera position of the selected image and the camera positions of the other images, and then we collect $k$-nearest neighbors as the data owned by a client\footnote{In our envisioned scenario, the clients are basically drones and vehicles, and the collected data should be sequential frames. Thus, we generate the client data from images that are physically close to each other.}.
We repeat this procedure for the number of clients, $N$, that is set to 100 in the experiments.
In this experiment, we randomly set $k$ for each client in a range of 100 to 200.
After training $N$ local models, we update the global model with the pose alignment and evaluate it with the validation data of Mill19.
We set the voxel size for caching the output of the local model to 0.25~m.
Following \cite{mega-nerf}, we use the appearance vector to model the color of each image.
For evaluation, we use the left half of the evaluation images for training, and evaluate the models on the right half of each image, as in \cite{mega-nerf}.
The experiments were conducted using the NVIDIA V100 GPU.

We use InstantNGP~\cite{ngp} as the local model because it converges faster and has fewer floating-point number operations than the original NeRF architecture, which is suitable if the computational resources of the clients are limited.
Each local model is trained for one epoch\footnote{One epoch corresponds to the iterations of \texttt{\#pixels} / \texttt{\#batchsize}.} with a batch size of 8192.
We optimize the local models with Adam~\cite{adam}, whose hyperparameters are the same as those of InstantNGP\cite{ngp}, except for the learning rate; it is set to 5e-3 only for the hash encoding and 5e-4 for the other parameters.

\subsection{Effect of Pose Alignment}
To evaluate our global pose alignment framework, we randomly select one local model as the global model.
We also randomly select another model as a local model from the models whose training data partially overlap the area modeled by the global model.
After selecting the models, we compute a center position of the overlapped area as a translation vector, $t^\ast\in\mathbb{R}^3$, and obtain rotation matrices, $\{R^\ast_j\in \mathrm{SO}(3)\}_j$, from camera poses of the training data close to $t^\ast$.
We render the images at $\{(t^\ast, R^\ast_j)\}_j$ as the target values through the global model.
Finally, we randomly sample a translation vector, $t\in\mathbb{R}^3$, and a rotation matrix, $R\in \mathrm{SO}(3)$, which can be regarded as the sensor noises, and optimize them by solving eq. \eqref{eq:pose-align}.
Thus, the optimal values will be $t=(0,0,0)$ and $R=I$, and we report the gap from them.
Note that since we use the appearance vector, as described in \ref{sec:setup}, we also optimize it in addition to the pose.
For optimization, we use the Adam optimizer~\cite{adam} with a batch size of 4096.
We set an initial learning rate to 5e-4 and decay it to 5e-5.

As an ablation study, we show the rotation errors and translation errors with various $\lambda$ in eq. \eqref{eq:pose-align} and a various number of target views in Tab. \ref{tab:ablation-pose}.
We randomly sample noises in a range of [-20~m, 20~m] and [-20$^{\circ}$, 20$^{\circ}$].
\begin{table}[]
    \caption{Rotation and translation errors with various $\lambda$ and a various number of target views. We report the mean and standard deviation over 10 trials.}
    \label{tab:ablation-pose}
    \begin{subtable}[t]{0.45\linewidth}
    \centering
        \begin{tabular}[t]{c|cc}
            $\lambda$ &  Rot. (deg) & Trans. (m)\\ \hline
            0.0  & 24.3$\pm$15.5 & 32.3$\pm$18.7\\
            0.25 & 19.9$\pm$16.7 & 17.7$\pm$11.2\\
            0.5  & 12.9$\pm$16.3 & 9.62$\pm$11.7\\
            0.75 & \textbf{0.53$\pm$0.96} & \textbf{0.19$\pm$0.33}\\
            1.0  & 0.92$\pm$1.63 & 0.43$\pm$0.78
        \end{tabular}
    \end{subtable}\hspace{4mm}
    \begin{subtable}[t]{0.45\linewidth}
    \centering
        \begin{tabular}[t]{c|cc}
            \#views &  Rot. (deg) & Trans. (m)\\ \hline
            1 & 0.34$\pm$0.24 & 0.97$\pm$0.68\\
            2 & 0.33$\pm$0.24 & 0.49$\pm$1.18\\
            4 & 0.53$\pm$0.96 & 0.19$\pm$0.33 \\
            8 & \textbf{0.12$\pm$0.09} & \textbf{0.02$\pm$0.01}
        \end{tabular}
    \end{subtable}
\end{table}
As presented on the left-hand side of Tab. \ref{tab:ablation-pose}, RGB information is critical for the pose alignment, but the depth information helps marginally with error reduction.
The multiple target views also contribute to reducing errors, as depicted on the right-hand side of Tab. \ref{tab:ablation-pose}.

We also show the evaluation results of our alignment framework with various magnitudes of initial errors in Fig. \ref{fig:var-noise}.
We randomly sample the pose noises and then align them.
We plot the results for 100 trials in Fig. \ref{fig:var-noise}.
Our method can align the translation error up to 75~m, which is comparable to a typical GPS error.
Also, it can correct the rotation error that is larger than errors in a typical IMU.
A few plots in Fig. \ref{fig:var-noise} indicate relatively large errors after alignment, but we can decrease the errors by increasing the number of particles and resampling rounds in the Monte Carlo-based optimization; in fact, our method can reduce errors even when the initial errors are larger than that of the failure cases.
In our method, the increase in computation time is acceptable because alignment will be performed offline and does not require real-time processing.

\begin{figure*}
    \centering
    \begin{tabular}{ccc}
       \includegraphics[clip,width=0.3\hsize]{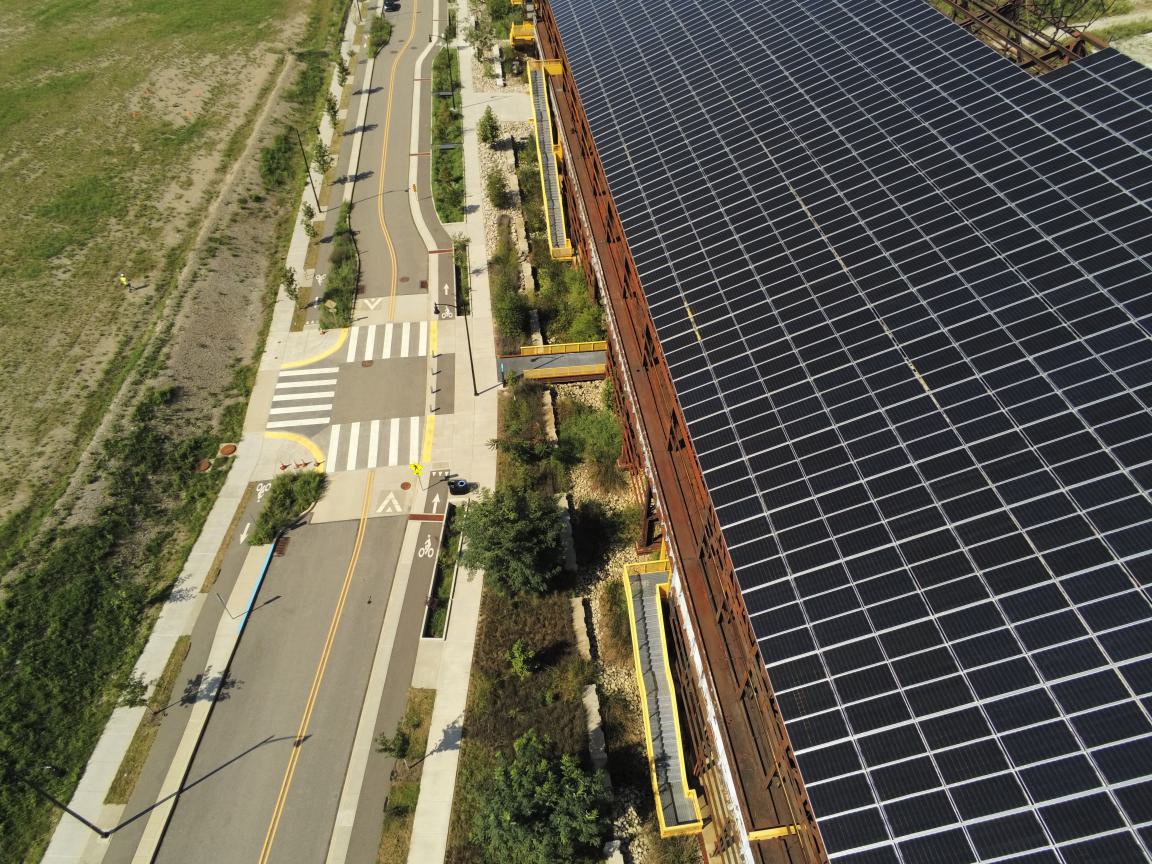} & 
       \includegraphics[clip,width=0.3\hsize]{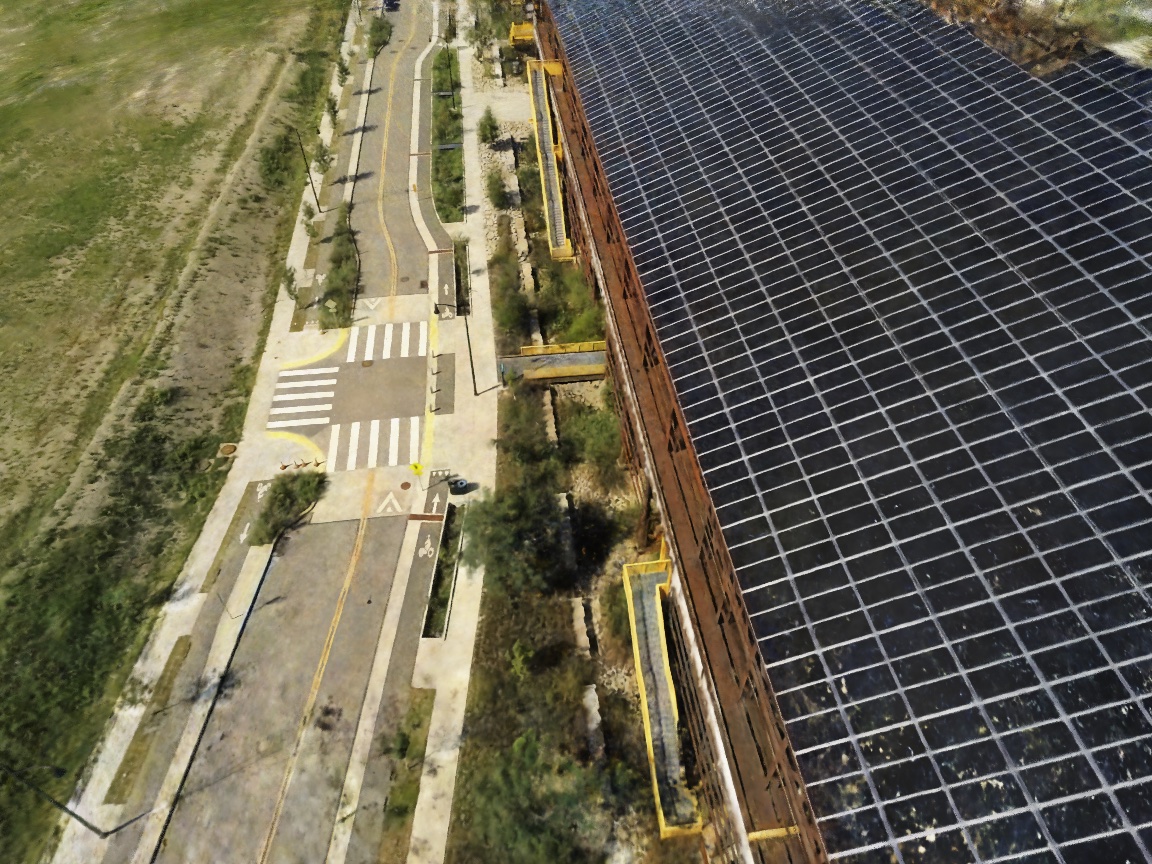} &
       \includegraphics[clip,width=0.3\hsize]{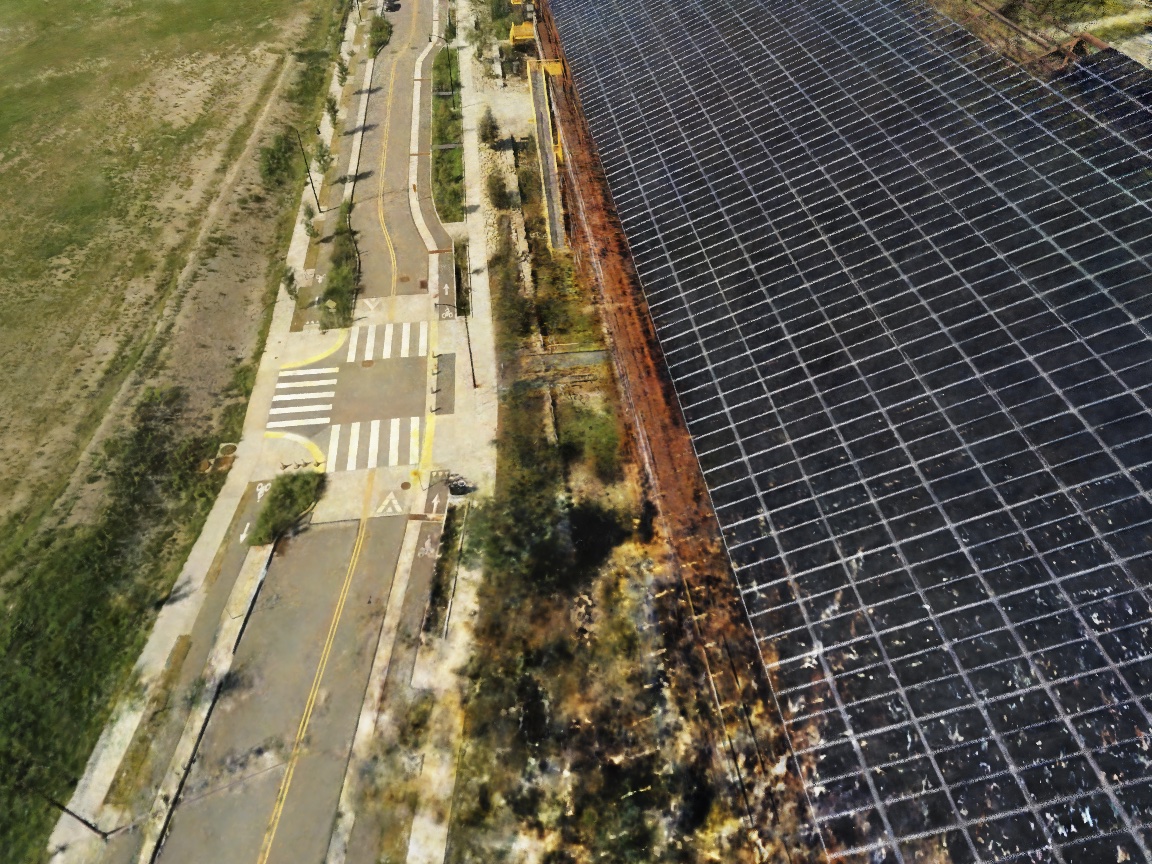}
    \end{tabular}
    \caption{From left to right, the ground-truth image of the test view, the image rendered by the local model trained with a sufficiently large number of viewpoints around the test view, and the image rendered by the local model trained with a relatively small number of viewpoints.}
    \label{fig:rendered-images}
\end{figure*}

Note that a few trials with various translation errors have failed, and the failure rate is 5\%.
It is due to the repetitive pattern in the scene.
As shown in Fig. \ref{fig:rendered-images}, there are grid patterns, and the alignment is failure when these patterns occupy the majority of the target image.
\begin{table*}[]
\centering
\caption{Performance for different training protocols on building and rubble scenes of Mill19.}
\label{tab:comparison}
\begin{tabular}{c|cc|cc|c}
                      & \multicolumn{2}{c|}{Building} & \multicolumn{2}{c|}{Rubble}   & Rendering Speed \\
                      & PSNR     & Training Time (h)  & PSNR    & Training Time (h)   & (Pixel/Second) \\ \hline
Baseline              & 20.24    & 32.4               & 22.01   &     28.0            &   80.4K        \\
Distributed Training  & 18.43    & 0.75               & 19.93   &     0.67            &   69.2K        \\
Ours                  & 17.51    & 0.97               & 20.12   &     0.95            &   342.6K  
\end{tabular}
\end{table*}
\begin{figure}
    \centering
    \begin{tabular}{cc}
       \includegraphics[clip,width=0.45\hsize]{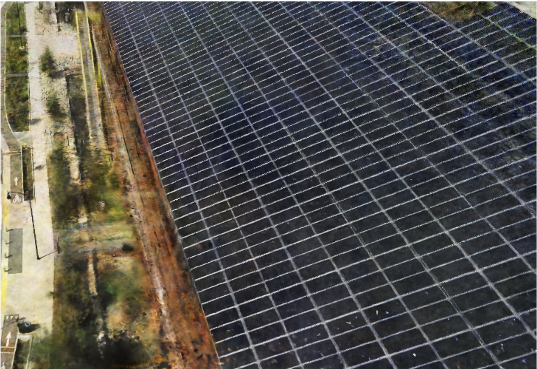} & 
       \includegraphics[clip,width=0.45\hsize]{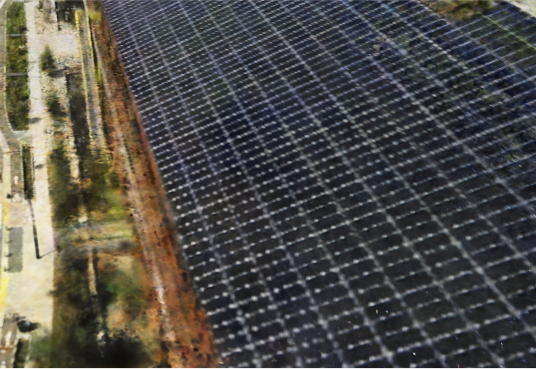}
    \end{tabular}
    \caption{An image rendered by the client model (left) and an image rendered by the cached voxel grids (right). The lattice pattern composed of high-frequency components is broken.}
    \label{fig:quant-err}
\end{figure}

\subsection{Comparison with Other Training Protocols}
\label{sec:comparison}
To evaluate the rendered image quality of our method, we compare the proposed training pipeline with two data-centralized training protocols, baseline and distributed training.
The baseline indicates training one model with all data, which corresponds to the ordinary NeRF training pipeline.
The distributed training indicates the Mega-NeRF's training pipeline~\cite{mega-nerf}; namely, the scene is divided into grids, and then models are trained with data corresponding to each grid.
We divide the scene into 4$\times$4 grids for the distributed training; that is, 16 models are trained, and the number of data for training each model is approximately 150, which is the same as the expected number of data used to train each model of our pipeline.
For a fair comparison, we evaluate these training pipelines with the same model and the same hyperparameters as ours, except for the model size in the baseline.
In other words, unlike Mega-NeRF, our distributed training do not use appearance modeling, and the number of training iterations is less.
Note that we assume that the global position is correctly aligned by our position alignment algorithm in this experiment.

We show the evaluation results in Tab. \ref{tab:comparison}.
Note that since the models can be trained in parallel, the longest training time out of all local models is reported as the training time in the distributed learning and ours.
The training time of distributed training and ours is much shorter than that of the baseline because of the distributed training protocol.
Since the scale of Mill19 is relatively small compared to our envisioned scenario, the training time of the baseline is acceptable.
However, if we scale it to an earth-scale, it is impossible to train a model with the standard training pipeline.
In addition to the reduction in training time, our method can alleviate bandwidth consumption because the size of the local model is 0.1~GB while that of the local data is up to 1~GB.
The rendering speed for the distributed training is slower than the others because the distributed training combines the outputs of the multiple models to render one image.
Rendering by ours is the fastest because it only samples cached outputs from the voxel grids for rendering, unlike the others that consist of hash encoding and MLP.

The PSNR for our method is worse than that for the baseline for two reasons: each client trains a local model with a relatively small number of data, while the baseline trains the model with a sufficiently large number of data.
Basically, NeRF requires a sufficiently large number of viewpoints to correctly represent the scene in 3D.
However, some clients do not satisfy this requirement around the test view and the local models trained by such clients cannot represent the scene correctly, as shown in Fig. \ref{fig:rendered-images}.
Consequently, such local models degrade the global model performance.
This disadvantage is also found in distributed training; in fact, PSNR for distributed training is worse than that for the baseline.
The other reason is the quantization error in the caching process.
We cache the outputs of the local models on the voxel grid in the aggregation step, and such a quantization operation approximates the continuous function by the piece-wise linear function.
Thus, there are approximation errors, which cause the gap between PSNR for distributed training and ours, especially for the building scene that includes high-frequency components, as shown in Fig. \ref{fig:quant-err}.
This error can be reduced by increasing the cached grid size, but it requires larger memory budgets.
Improving local model training and alleviating the quantization errors are possible directions for future work.

The PSNR for ours on the rubble scene is better than that for distributed training.
The quantization errors in the rubble scene are smaller than those in the building scene because images in the rubble scenes are basically composed of low-frequency components.
Therefore, in scenes where the effect of the errors is small, our training protocol would be better than distributed training in this experimental setting.

\section{Limitations and Future Work}
As we mentioned in Sec. \ref{sec:assumption}, this work does not consider dynamic and/or transient objects and lighting changes.
It may raise privacy concerns because if a person in the scene is modeled with the lighting, the model may show where and when the person was doing what.
However, we can remove such objects using image segmentation models~\cite{suzuki2022clustering,cheng2022masked} as in Block-NeRF~\cite{block-nerf} and can model multiple appearances using an appearance vector as in NeRF-W~\cite{nerf-w}; by exploiting the advantage of such methods, the proposed pipeline will leverage the privacy-preserving aspect of the federated learning.
Therefore, we believe that considering the transient objects and lighting changes in the federated learning pipeline is an important research direction.

Moreover, as we mentioned in Sec. \ref{sec:comparison}, the quality of the rendered images is worse than the images rendered by the baseline.
Fortunately, there are many studies on training NeRF with limited viewpoints~\cite{deng2022depth,yu2021pixelnerf,kim2022infonerf}.
We believe that we can improve the performance of the local models by leveraging the training protocol of such methods; consequently, the global model can be improved.
Another approach is to consider the number of viewpoints for aggregating models.
Specifically, by adjusting the mixing coefficient in the exponential moving average according to the number of viewpoints, we may be able to ignore the low-quality outputs of the local model in the aggregation step.

\section{Conclusion}
In this study, we proposed the federated learning pipeline for large-scale scene modeling with NeRF.
We designed the aggregation step for training NeRF in a federated learning manner: The global model is defined as the voxel grids and the local models are aggregated by caching the outputs of the local model on the voxel grids, which allows us to update the global model locally.
In addition, we proposed the global pose alignment step to alleviate the sensor noise in the global pose of each client.
We assessed the proposed method on the Mill19 dataset containing thousands of images collected from drone footage over 100,000m$^2$ of terrain near an industrial complex and verified its effectiveness.

We believe our work opens new avenues for large-scale scene modeling using NeRF and is an important step towards collaboratively training for NeRF in a federated learning manner.





\bibliographystyle{IEEEtran}
\bibliography{reference}

\end{document}